\documentclass{article}
\usepackage{PRIMEarxiv}
\usepackage{microtype}
\usepackage{graphicx}
\usepackage{subcaption}
\usepackage{booktabs} 
\usepackage{balance}

\usepackage{hyperref}

\usepackage{amsmath}
\usepackage{amsfonts}
\usepackage{amssymb}
\usepackage{mathtools}
\usepackage{amsthm}
\usepackage{multirow}
\usepackage{pifont}
\usepackage{wrapfig}

\usepackage[capitalize,noabbrev]{cleveref}

\theoremstyle{plain}

\theoremstyle{definition}

\theoremstyle{remark}

\usepackage[textsize=tiny]{todonotes}

\begin{document}

\title{GRAMformer: Any-Order Modality Interactions\\via Volumetric Multimodal Cross-Attention}
\author{Giordano Cicchetti, Eleonora Grassucci, Danilo Comminiello\\Dept. of Information Engineering, Electronics, and Telecommunications, Sapienza University of Rome, Italy\\\{name.surname\}@uniroma1.it}
\maketitle

\begin{abstract}
Transformer-based multimodal models rely on attention mechanisms to integrate information across heterogeneous modalities. Despite their success, existing multimodal attention formulations compute their scores through collections of pairwise dot-product interactions or by concatenating all the modalities into the keys, even when multiple modalities should be jointly involved. As a consequence, current approaches either incur quadratic complexity in the number of modalities or fail to explicitly model interactions that depend on the joint configuration of multiple representations.
In this work, we introduce the Volumetric Multimodal cross-Attention (VMA), a novel cross-attention mechanism in which attention scores are defined as a function of the joint geometry of a query and multiple modality-specific keys.
VMA computes the volume spanned by query and key vectors across multiple modalities, capturing joint multimodal dependencies beyond pairwise similarity, enabling native modeling of any-order modality interactions.
We integrate VMA into our novel multimodal transformer architecture, named GRAMformer, explicitly designed to integrate any number of modalities. We evaluate the proposed model on multimodal learning tasks, demonstrating improved effectiveness and efficiency.
Code available at \url{https://github.com/ispamm/GRAMformer/tree/main}.
\end{abstract}


\section{Introduction} 

\begin{wrapfigure}{r}{0.5\textwidth}
    \vspace{-0.60cm}
    \centering
    \includegraphics[width=\linewidth]{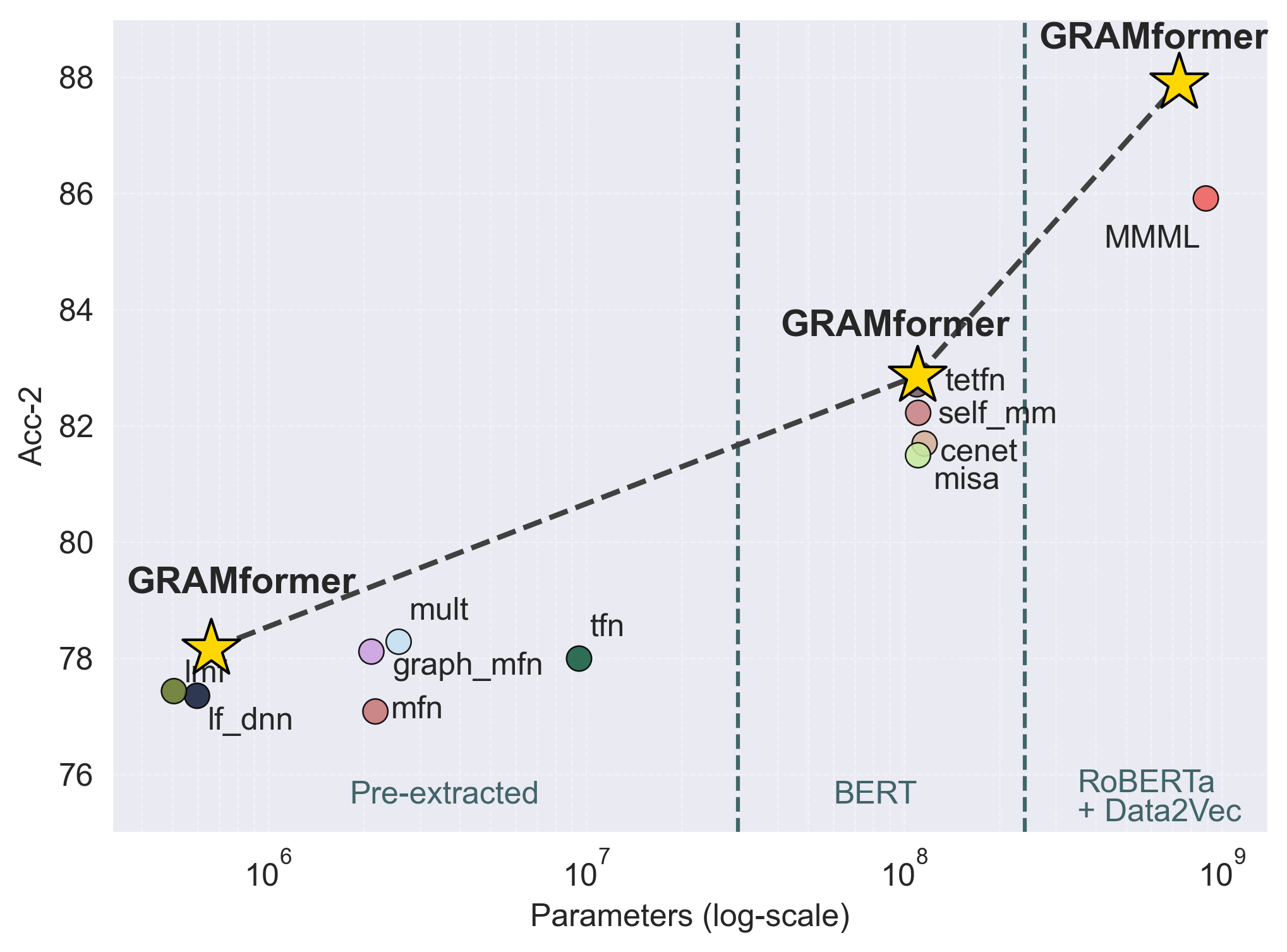}
    \caption{Acc-2 in the MOSI dataset under different encoders. Our GRAMformer outperforms comparisons while being lighter.}
    \label{fig:fig1}
    \vspace{-0.30cm}
\end{wrapfigure}
Attention-based transformer architectures \cite{Vaswani2017NIPS} are at the core of most modern unimodal and multimodal learning systems \cite{dosovitskiy2021ViT, Alayrac2022Flamingo, Ye2024mPLUGOwl3TL}. In multimodal transformers, the cross-attention mechanism is responsible for integrating information across heterogeneous modalities within a shared representation space \cite{Lu2019vilbert}. However, while highly effective in unimodal and bimodal settings, the attention operation itself is inherently pairwise, evaluating relevance independently between a query and each modality-specific key. As a result, current multimodal transformers still model interactions across modalities through collections of pairwise similarities \cite{Tsai2019MUlT, esser2024scaling, lv2025taca}, even when attention relevance depends on the joint agreement of multiple modalities.

\begin{figure*}
    \centering
    \includegraphics[width=\linewidth]{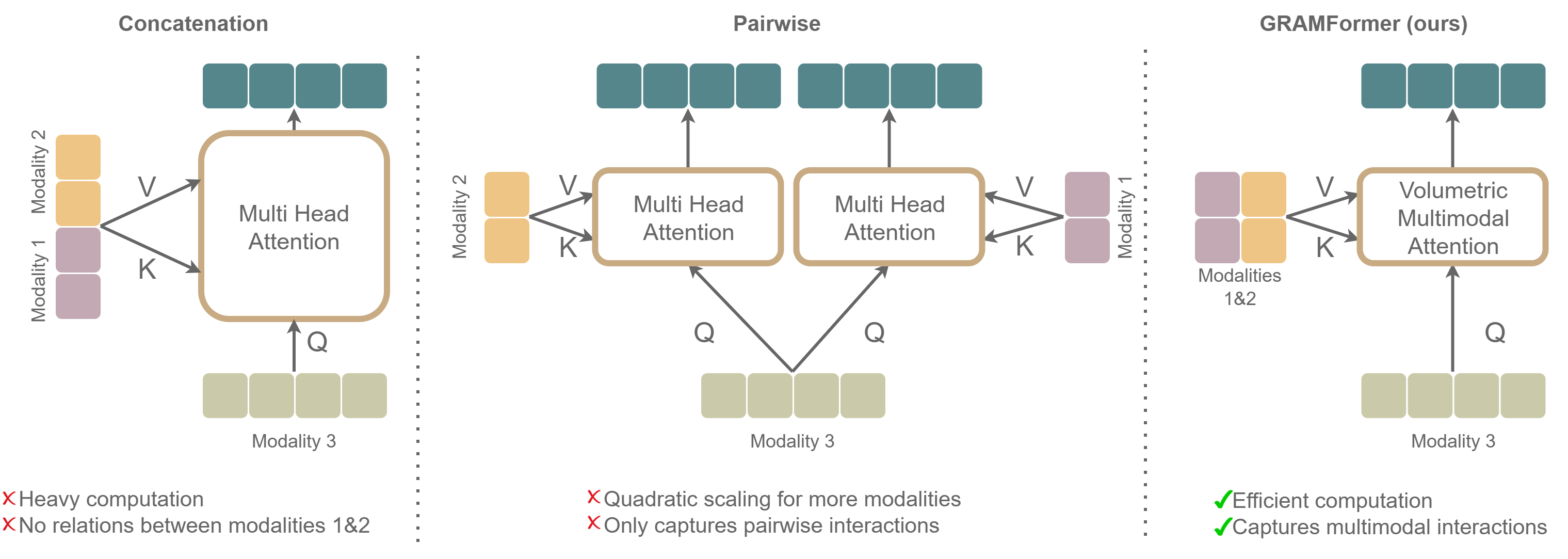}
    \caption{Comparison among conventional multimodal cross-attention mechanisms (concatenation, pairwise dot product) and proposed Volumetric Multimodal Attention capable of computing joint attention scores among all the modalities.}
    \label{fig:MHA_comparisons}
\end{figure*}

Existing multimodal fusion strategies address this limitation only indirectly.
On one side, Early-fusion transformers and multimodal large language models (MLLMs) either concatenate all modality tokens for joint self-attention \cite{Sun2019ICCV, Mo2024CVPR, huang2023language} or inject modality features into language decoders via cross-attention adapters \cite{zhang2023VideoLLaMA, Alayrac2022Flamingo, Li2023BLIP2}. These approaches suffer from high memory and compute demands as the key and value stack dramatically grows. Additionally, no interactions are computed between concatenated modalities in keys and values, limiting the expressiveness of the resulting representation. 
On the other side, pairwise cross-attention mechanisms \cite{Tsai2019MUlT, Zhao2025SciRep} compute relevance independently for each modality and aggregate the results, leading to quadratic scaling in the number of modalities and failing to capture higher-order modal dependencies.
Recent extensions toward tri-modal or tensorized attention \cite{Zhou2021TripleAtt, ijcai2022p480, Nie2024TripletAT} approximate interactions through factorization or triplets, but still decompose relevance into lower-order components or require severe approximations to contain computational demands. Consequently, none of these approaches explicitly defines attention as a function over a set of modality representations.

Nevertheless, modeling the interactions among multiple modalities pairwisely via dot product fails to capture joint semantic information and whether a set of representations lies in a common low-dimensional subspace or spans a higher-dimensional region of the embedding space. This distinction can be quantified through the volume of the parallelotope spanned by the representations, which is low when the vectors are close to each other and increases as their joint span grows \cite{cicchetti2025gram}. As such, volume provides a scalar measure of joint linear dependence among a set of embeddings, making it a suitable candidate for defining attention scores that simultaneously depend on multiple modality representations.
Building on this insight, we propose Volumetric Multimodal cross-Attention (VMA), a novel attention mechanism in which relevance is computed from the joint geometry of a query and multiple modality-specific keys. Instead of evaluating similarity pairwise, VMA defines attention logits using the volume of the parallelotope spanned by the query and the set of modality keys. This formulation enables attention to respond directly to higher-order multimodal interactions while efficiently scaling with any number of modalities. We integrate VMA into the proposed GRAMformer, our novel transformer architecture for multimodal information fusion. Through VMA, the GRAMformer exploits multimodal interactions effectively without requiring heavy pairwise computations or concatenation operations, yet jointly modeling multimodal attention scores. This results in a lightweight architecture with improved performance in any tested downstream task.



Our main contributions can be summarized as follows:
\begin{itemize}
    \item We identify a fundamental limitation of pairwise attention mechanisms in modeling higher-order multimodal dependencies.
    \item We introduce Volumetric Multimodal cross-Attention (VMA), a novel cross-attention mechanism that enables joint attention scores defined over any-order modality representations without relying solely on pairwise computations.
    \item We integrate VMA into a novel transformer architecture, the GRAMformer, demonstrating its effectiveness and efficiency across multimodal tasks.
\end{itemize}

\section{Related Work} 

\textbf{Multimodal Learning.} Since CLIP \cite{Radford2021LearningTV}, the multimodal representation learning research field has seen a growing interest. Connecting two modalities, such as text and audio \cite{CLAP2022} or video and text \cite{Luo2021CLIP4ClipAE} in the latent space has become crucial for downstream tasks ranging from retrieval \cite{Girdhar2023ImageBindOE, Wang2024InternVideo2SV} to generation \cite{liu2025flowing}. More recently, several studies moved towards defining a unified multimodal representation for more than two modalities. Among them, TRIANGLE \cite{cicchetti2025triangle} aligns triplets by minimizing the area of a triangle in the latent space, Symile \cite{saporta2024contrasting} by working with the total correlation among multiple modalities, GRAM \cite{cicchetti2025gram} by leveraging the volume of the parallelotope spanned by $n$-modality embeddings, and PRML \cite{liu2025principledmultimodalrepresentationlearning} working on volume eigenvalues.

\textbf{Cross-attention.} Existing multimodal attention methods can be broadly divided into two families. Pairwise similarity–based approaches compute all bi-modal attentions \cite{Tsai2019MUlT, Zhao2025SciRep}. Therefore, in the case of three modalities, such methods repeat the attention computation six times (two directions, three modalities). This formulation scales quadratically with the number of modalities requiring heavy computations and only captures pairwise interactions, lacking a unified control.
Another family of methods relies on concatenating the tokens from all the modalities together to perform a joint cross-attention with a query modality \cite{zhang2023VideoLLaMA, Alayrac2022Flamingo, Li2023BLIP2, Lu2019vilbert}. Nevertheless, these methods solely compute the attention scores between the stacked modalities and the query one, with no clues on the scores among the stacked ones. Moreover, the computational demand dramatically grows as the number of modalities increases.

\textbf{Higher-order attention.} Recently, some works tried to scale the attention computation to three modalities, relying on triplet products or tensorized representations \cite{Zhou2021TripleAtt, ijcai2022p480, Nie2024TripletAT, rao2025quadruple}.
However, their expressivity is constrained by rank selection \cite{ijcai2022p480} and approximations due to computational infeasibilities \cite{Zhou2021TripleAtt, rao2025quadruple}, which can suppress fine-grained or modality-specific dependencies. 
In contrast, and differently from previous multimodal learning loss-oriented works, the proposed Volumetric Multimodal Attention augments the dot-product similarity with a volume-based measure directly in the attention score computation, enabling native modeling of true multimodal interactions without pairwise decomposition or restrictive low-rank assumptions.

\section{Volumetric Multimodal Cross-Attention} 

\subsection{Preliminaries}



\begin{wrapfigure}{r}{0.5\textwidth}
    \vspace{-0.60cm}
    \centering
    \includegraphics[width=\linewidth]{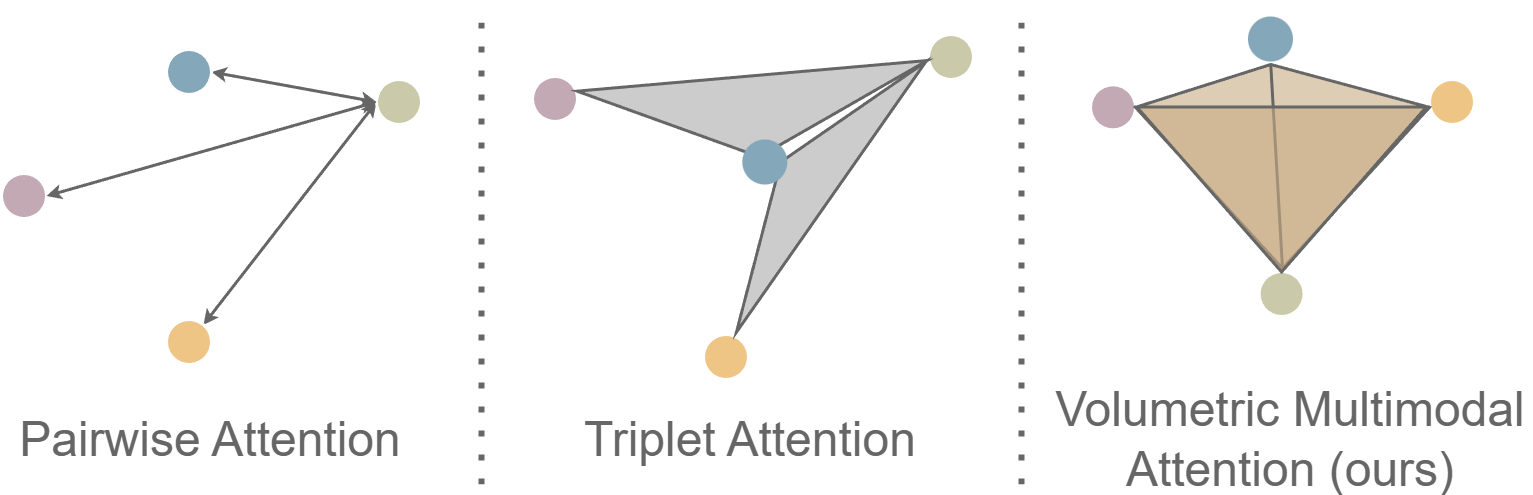}
    \caption{Pairwise, triplet and Volumetric Multimodal Attention (VMA) schematic representations.}
    \label{fig:att}
\end{wrapfigure}

\paragraph{Multihead Cross Attention.}
Multihead Cross-Attention is a core mechanism for modeling interactions between two distinct sets of representations, namely queries and keys-values, and it is at the heart of common multimodal transformers for cross-modal conditioning \cite{esser2024scaling, lv2025taca}. 
Let $\mathbf{Q} \in \mathbb{R}^{N_q \times d}$ denote a sequence of query embeddings and $\mathbf{K}, \mathbf{V} \in \mathbb{R}^{N_k \times d}$ denote the corresponding key and value embeddings, potentially originating from a different modality or representation space.
Cross-attention computes, for each query, a weighted combination of the values based on its similarity to the keys in the following way:
\begin{equation}
\mathrm{Attn}(\mathbf{Q},\mathbf{K},\mathbf{V})
= \mathrm{softmax}\!\left(
\frac{\langle \mathbf{Q},  \mathbf{K} \rangle}{\sqrt{d}}
\right)\mathbf{V}
\label{eq:crossatt_normal}
\end{equation}

In the multihead formulation, the input representations are first linearly projected into $H$ parallel subspaces:
\begin{equation}
\mathbf{Q}^h = \mathbf{Q}\mathbf{W}^h_Q, \quad
\mathbf{K}^h = \mathbf{K}\mathbf{W}^h_K, \quad
\mathbf{V}^h = \mathbf{V}\mathbf{W}^h_V,
\end{equation}
where $\mathbf{W}^h_Q, \mathbf{W}^h_K, \mathbf{W}^h_V \in \mathbb{R}^{d \times d_h}$ are learned projection matrices and $d_h = d / H$ is the dimensionality of each head.

Each attention head computes scaled dot-product attention independently:
\begin{equation}
\mathrm{Attn}_h(\mathbf{Q}_h,\mathbf{K}_h,\mathbf{V}_h)
= \mathrm{softmax}\!\left(
\frac{\langle \mathbf{Q}_h,  \mathbf{K}_h \rangle}{\sqrt{d_h}}
\right)\mathbf{V}_h,
\label{eq:crossatt}
\end{equation}
with $\langle \cdot,  \cdot \rangle$ dot product, producing a head-specific output that captures a particular pattern of cross-modal interaction.
The outputs of all heads are then concatenated and linearly transformed to produce the final cross-attention representation:
\begin{equation}
\mathrm{MultiHead}(\mathbf{Q},\mathbf{K},\mathbf{V})
= \mathrm{Concat}\big(\mathrm{Attn}_1,\dots,\mathrm{Attn}_H\big)\mathbf{W}^O,
\end{equation}
where $\mathbf{W}^O \in \mathbb{R}^{d \times d}$ is a learned output projection.

By attending to multiple subspaces in parallel, multihead cross-attention enables the model to capture diverse and complementary relationships between heterogeneous representations, which is particularly important in multimodal learning scenarios.

\paragraph{Limitations in multimodal scenarios.} 
\label{sec:limitations}
Conventional cross-attention relies on scaled dot-product similarity to compute attention weights between a set of query tokens and a set of key-value tokens as in \eqref{eq:crossatt}.
This formulation captures pairwise interactions between individual query and key tokens and has proven highly effective when the key-value representations originate from a single modality. However, extending cross-attention to settings where multiple conditioning modalities are involved is non-trivial. When key-value tokens originate from multiple modalities, standard dot-product attention lacks the ability to explicitly model higher-order cross-modal interactions. Attempts to overcome this crucial limitation have concentrated on two strategies.

\textit{1. Token concatenation across modalities.}
A common approach is to concatenate the key-value tokens from multiple modalities into a single sequence \cite{zhang2023VideoLLaMA}. Given two conditioning modalities with key representations $\mathbf{K}_1 \in \mathbb{R}^{N_1 \times d}$ and $\mathbf{K}_2 \in \mathbb{R}^{N_2 \times d}$, one can define
\begin{equation}
\mathbf{K} = [\mathbf{K}_1; \mathbf{K}_2], \quad
\mathbf{V} = [\mathbf{V}_1; \mathbf{V}_2],
\end{equation}
with $\mathbf{K},\mathbf{V} \in \mathbb{R}^{(N_1+N_2)  \times d}$.

In this setting, each query token independently attends to individual tokens from either modality through pairwise dot products. While simple and efficient, this approach treats tokens from different modalities as independent entities. As a consequence, the attention mechanism cannot jointly reason over aligned multimodal tokens, even when the modalities are temporally or semantically synchronized. Interactions between modalities are only implicit and mediated through the query, resulting in a lack of true multimodal, higher-order token interactions.

\textit{2. Pairwise cross-attention architectures.}
An alternative strategy consists of modeling cross-modal interactions using separate cross-attention modules for each modality pair \cite{Tsai2019MUlT}. A query modality may attend independently to two conditioning modalities via two distinct transformers as:
\begin{equation}
\mathrm{Attn}_1(\mathbf{Q}, \mathbf{K}_1, \mathbf{V}_1), \quad
\mathrm{Attn}_2(\mathbf{Q}, \mathbf{K}_2, \mathbf{V}_2).
\end{equation}
Although this design allows a query modality to grasp correlations  from each modality separately, repeating this process for each modality it scales quadratically with the number of modalities in terms of parameters and memory. More importantly, it still fails to capture higher-order interactions involving more than two modalities simultaneously. Cross-modal reasoning remains restricted to pairwise relationships, preventing the model from explicitly representing joint multimodal alignment, as shown in Fig.~\ref{fig:att}.

\textit{Discussion.}
In both cases, standard cross-attention mechanisms are fundamentally limited by their reliance on pairwise similarity measures. They lack an explicit mechanism to model joint interactions among multiple conditioning modalities, which is particularly problematic in settings where modalities are aligned or mutually informative, like video-audio-text scenarios. These limitations motivate the need for attention mechanisms that can directly capture higher-order multimodal relationships while preserving the efficiency and inductive biases of standard cross-attention.

\subsection{Proposed Volumetric Multimodal Cross-Attention}
\begin{wrapfigure}{r}{0.5\textwidth}
    \vspace{-0.60cm}
    \centering
    \includegraphics[width=\linewidth]{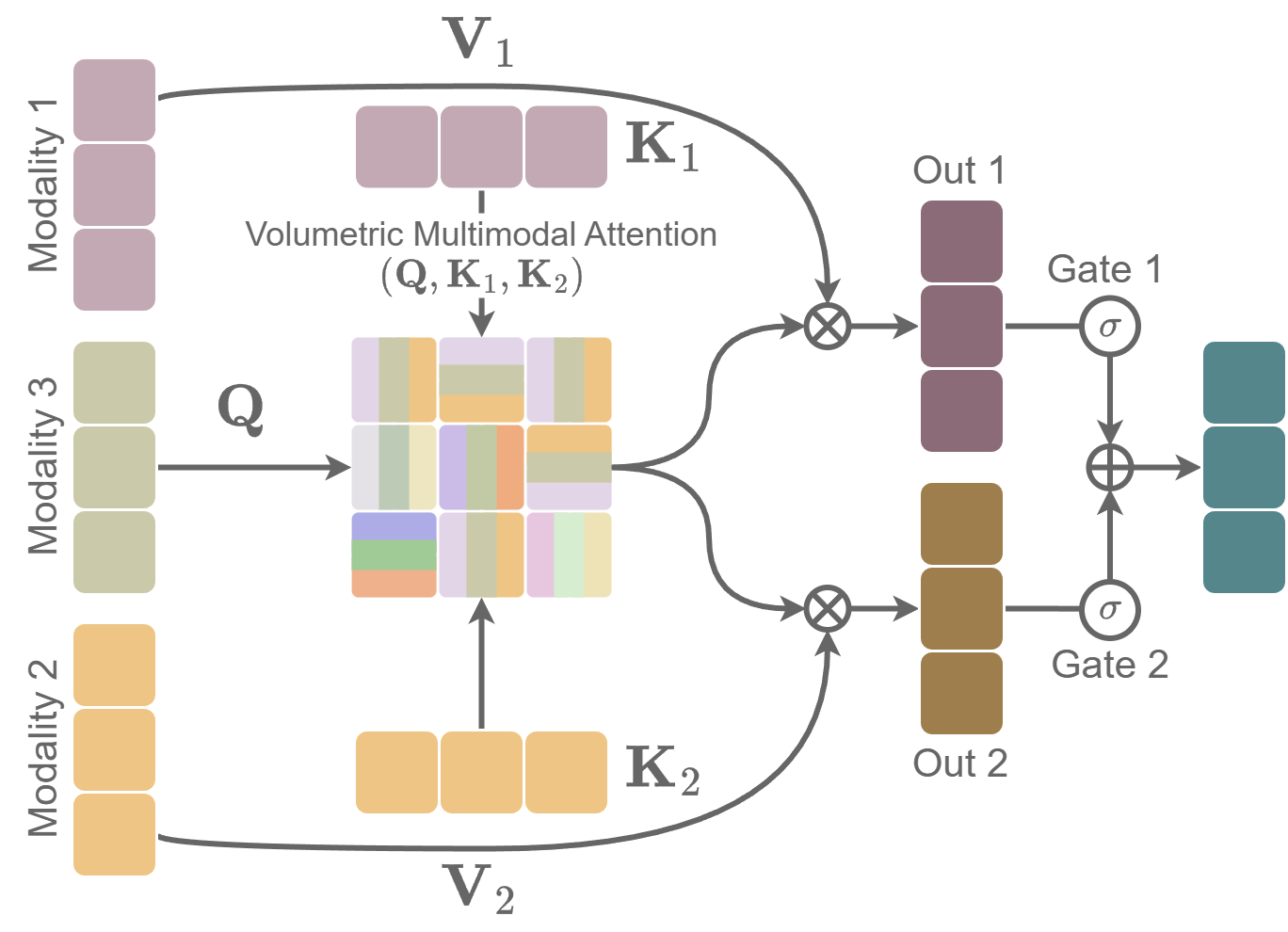}
    \caption{Proposed Volumetric Multimodal cross-Attention for three modalities. Importantly, the scheme can be extended up to $d$ modalities, with $d$ embedding dimension.}
    \label{fig:method}
    \vspace{-0.50cm}
\end{wrapfigure}

In this paper, we propose the novel Volumetric Multimodal cross-Attention (VMA) mechanism that jointly computes attention scores across $M$ conditioning modalities while remaining efficient and lightweight. The key idea is to replace purely pairwise similarity with a unified geometric measure that captures higher-order multimodal alignment, while preserving the inductive biases of standard attention.

\textbf{Setting.}
Let us consider a directed multimodal cross-attention setting in which a query modality attends jointly to $M$ heterogeneous conditioning modalities. Let $\mathbf{Q} \in \mathbb{R}^{N_q \times d}$ denote the query embeddings, and let $\{(\mathbf{K}_m, \mathbf{V}_m)\}_{m=1}^M$ denote the key-value embeddings of the conditioning modalities, with $\mathbf{K}_m, \mathbf{V}_m \in \mathbb{R}^{N_m \times d}$. Tokens across modalities are assumed to be aligned along the sequence dimension for design choice, such that the $j$-th token of each modality forms a multimodal key group. For the sake of simplicity, let us assume all modalities provide the same number of aligned tokens here.
%
%

\textbf{Volumetric attention score computation.}
Volumetric attention scores are computed between query tokens and aligned multimodal key groups. For a query token $\mathbf{q} \in \mathbb{R}^{d}$ and a set of key tokens $\{\mathbf{k}_1, \dots, \mathbf{k}_M\}$ drawn from the $M$ modalities at the same sequence index, we define the joint multimodal similarity score based on the volume spanned by these vectors.
Specifically, by arranging the vectors $[\mathbf{q}, \mathbf{k}_1, \dots, \mathbf{k}_M]$ as columns of a matrix $\mathbf{Z} \in \mathbb{R}^{d \times (M+1)}$, we construct the associated Gram matrix $\mathbf{G} = \mathbf{Z}^\top \mathbf{Z}$, with $\mathbf{G} \in \mathbb{R}^{(M+1)\times(M+1)}$. Then, the volume of the parallelotope spanned by the vectors is given by the square root of the determinant of $\mathbf{G}$ \cite{Gantmacher1959matrix} as:

\begin{equation}
    \mathrm{Vol}(\mathbf{q}, \mathbf{k}_1, \dots, \mathbf{k}_M) = \sqrt{\det(\mathbf{G}) + \epsilon},
\end{equation}

\noindent with $\epsilon$ a small constant value to ensure stability. This volume provides the geometric measure of joint multimodal alignment, as it is small when the representations are mutually aligned and increases as they become more independent \cite{cicchetti2025gram}.

Crucially, the volume remains meaningful for $M + 1 \le d$ \cite{Gantmacher1959matrix}, with typically $M + 1 \ll d$ in real-world scenarios as the embedding dimension $d$ is usually very large, enabling the VMA computation for any-order of modalities. Interestingly, such a measure is highly flexible as, for the minimal non-trivial case $M=2$, the volume reduces to the area of the parallelogram spanned by the two vectors, being zero if and only if the vectors coincide, while increasing as they span a higher-dimensional subspace. In addition, the volume is invariant under orthogonal transformations of the embedding space, as well as under permutations of the vectors, since these operations leave the Gram determinant unchanged \cite{Gantmacher1959matrix}. Consequently, the volumetric score does not depend on the ordering of the conditioning modalities and is agnostic to rotations of the representation space. 
To further preserve strong bimodal interactions,
we combine this higher-order geometric term with a dot-product regularization between the query and each modality, using a scaling factor $\beta$ to control the amount of high-order information injected into the attention mechanism. Therefore, the resulting attention logit is defined as:
\begin{equation}
\mathbf{s}(\mathbf{q}, \mathbf{k}_1, \dots, \mathbf{k}_M)
=  \\
= \frac{
- \ \beta  \ \mathrm{Vol}(\mathbf{q}, \mathbf{k}_1, \dots, \mathbf{k}_M)
+ \sum\limits_{m=1}^M \langle\mathbf{q},\mathbf{k}_m\rangle
}{
\sqrt{d}
}.
\label{eq:attscores}
\end{equation}

Considering all query tokens and all aligned multimodal key groups, we construct an attention score matrix $\mathbf{A} \in \mathbb{R}^{N_q \times N_m}$, where each entry $\mathbf{A}_{i,j}$ measures the relevance between the $i$-th query token $\mathbf{q}_i$ and the $j$-th multimodal key group $(\mathbf{k}_{1,j}, \dots, \mathbf{k}_{M,j})$:
\begin{equation}
\mathbf{A}_{i,j}
=
\frac{
- \ \beta  \ \mathrm{Vol}(\mathbf{q}_i, \mathbf{k}_{1,j}, \dots, \mathbf{k}_{M,j})
+ \sum\limits_{m=1}^{M} \langle \mathbf{q}_i, \mathbf{k}_{m,j} \rangle
}{
\sqrt{d}
}.
\end{equation}

\begin{table*}[t]
\centering
\caption{Comparison on MOSI and MOSEI Datasets. * represents results obtained from \cite{mao2022m} and its corresponding GitHub page. Models with † are reproduced under the same conditions. Best results are marked in bold. Results averaged over 5 runs.}
\label{tab:mosi_mosei}
\resizebox{\textwidth}{!}{
\begin{tabular}{l|cccccc|cccccc}
\toprule
\multirow{3}{*}{Method} & \multicolumn{6}{c|}{MOSI} & \multicolumn{6}{c}{MOSEI} \\
\cmidrule{2-13}
 & \multicolumn{2}{c}{Binary} & 5-class & 7-class & \multicolumn{2}{c|}{Regression}
 & \multicolumn{2}{c}{Binary} & 5-class & 7-class & \multicolumn{2}{c}{Regression} \\
 & Acc-2 $\uparrow$ & F1 $\uparrow$ & Acc-5 $\uparrow$ & Acc-7 $\uparrow$ & MAE $\downarrow$ & Corr $\uparrow$
 & Acc-2 $\uparrow$ & F1 $\uparrow$ & Acc-5 $\uparrow$ & Acc-7 $\uparrow$ & MAE $\downarrow$ & Corr $\uparrow$ \\
\midrule
TFN*      & 77.99/79.08 & 77.95/79.11 & -     & 34.46 & 0.947 & 0.673 & 78.50/81.89 & 78.96/81.74 & -     & 51.6  & 0.572 & 0.714 \\
LMF*      & 77.90/79.18 & 77.80/79.15 & -     & 33.82 & 0.950 & 0.651 & 80.54/83.48 & 80.94/83.36 & -     & 51.59 & 0.575 & 0.716 \\
MulT*     & 79.71/80.95 & 79.63/80.95 & 42.68 & 36.91 & 0.879 & 0.702 & 81.15/84.63 & 81.56/84.52 & 54.18 & 52.84 & 0.559 & 0.733 \\
ICCN      & -/83.07     & -/83.02     & -     & 39.01 & 0.862 & 0.714 & -/84.18     & -/84.15     & -     & 51.58 & 0.565 & 0.713 \\
MISA*     & 81.84/83.54 & 81.82/83.58 & 47.08 & 41.37 & 0.776 & 0.778 & 80.67/84.67 & 81.12/84.66 & 53.63 & 52.05 & 0.557 & 0.751 \\
MAG-BERT  & 82.37/84.43 & 82.50/84.61 & -     & 43.62 & 0.727 & 0.781 & 82.51/84.82 & 82.77/84.71 & -     & 52.67 & 0.543 & 0.755 \\
PMR       & -/82.40     & -/82.10     & -     & 40.60 & -     & -     & -/83.10     & -/82.80     & -     & 51.80 & -     & -     \\
MFSA      & -/83.3      & -/83.7      & -     & 41.1  & 0.856 & 0.722 & -/83.8      & -/83.6      & -     & 53.2  & 0.574 & 0.724 \\
FDMER     & -/84.6      & -/84.7      & -     & 44.1  & 0.724 & 0.788 & -/86.1      & -/85.8      & -     & 54.1  & 0.536 & 0.773 \\
Self-MM†  & 82.54/84.45 & 82.46/84.44 & 52.22 & 45.56 & 0.719 & 0.794 & 82.09/84.76 & 82.43/84.67 & 53.54 & 53.65 & 0.535 & 0.761 \\
ALMT†     & 83.24/85.37 & 83.41/85.46 & 50.29 & 44.75 & 0.738 & 0.776 & 82.34/84.95 & 81.85/85.93 & 55.05 & 53.32 & 0.534 & 0.771 \\
ConFEDE   & 84.17/85.52 & 84.13/85.52 & -     & 42.27 & 0.742 & 0.784 & 81.65/85.82 & 82.17/85.83 & -     & 54.86 & 0.522 & 0.780 \\
DEVA      & 84.40/86.29 & 84.48/86.30 & 51.78 & 46.32 & 0.730 & 0.787 & 83.26/86.13 & 82.93/86.21 & 55.32 & 52.26 & 0.541 & 0.769 \\

MMML & \underline{85.91}/88.16 & \underline{85.85}/88.15 & \underline{56.08} & 48.25 & \underline{0.643} & \underline{0.838}
     & 86.32/86.73 & 86.23/86.49 & \underline{57.32} & 54.95 & \underline{0.517} & \underline{0.790} \\

SPT & 81.20/- & 81.30/- & - & - & - & -
    & 82.40/- & 82.70/- & - & - & - & - \\

MMT & 85.80/87.00 & 85.80/87.00 & - & - & 0.657 & 0.830
    & - & - & - & - & - & - \\

FDMA & -/\underline{88.39} & -/\underline{88.36} & - & \underline{48.52} & 0.686 & 0.822
     & -/\textbf{87.41} & -/\textbf{87.38} & - & \underline{55.30} & \underline{0.517} & 0.782 \\

\midrule
GRAMformer
 & \textbf{87.90/89.94} & \textbf{87.87/89.94} & \textbf{57.87} & \textbf{50.05} & \textbf{0.593} & \textbf{0.871}
 & \underline{84.07}/\underline{87.39} & \underline{84.33}/\underline{87.26} & \textbf{58.07} & \textbf{56.05} & \textbf{0.4963} & \textbf{0.810} \\
\bottomrule
\end{tabular}
}
\end{table*}

Following standard cross-attention, softmax normalization is applied independently for each query token by normalizing each row of the attention score matrix $\mathbf{A}$. This produces, for each query token, a probability distribution over the aligned multimodal key groups, indicating how strongly the query attends to each joint cross-modal representation.

\textbf{Value aggregation.}
In our VMA, each conditioning modality provides a set of value tokens $\mathbf{V}_m$. We compute modality-specific attended values by multiplying the attention weights with the corresponding value representations. To dynamically control the contribution of each modality, we apply a query-dependent gating mechanism that modulates the extracted information from each modality, following \cite{qiu2025gated}.
The gated modality-specific values are then averaged to form the final multimodal output of the attention layer:
\begin{equation}
\text{VMA}(\mathbf{Q},\mathbf{K}_1,\dots,\mathbf{K}_M,\mathbf{V}_1,\dots,\mathbf{V}_M)
= \\
= \frac{1}{M} \sum_{m=1}^M
\mathbf{A} \mathbf{V}_m^{\top}
\odot
\sigma(\mathbf{W}_{\mathrm{gate},m} \mathbf{Q}),
\label{eq:gating}
\end{equation}

where $\mathbf{W}_{\mathrm{gate},m} \in \mathbb{R}^{d \times d}$ are learned gating matrices.
Figure~\ref{fig:method} shows the schematic representation of the proposed VMA mechanism.


Finally, the proposed Volumetric Multimodal cross-Attention can be naturally extended to the multi-head formulation; namely Multihead Volumetric Multimodal Cross-Attention (M-VMA). Given an embedding dimension $d$ and a number of heads $H$, the query, key, and value representations are projected into $H$ subspaces of dimension $d_h = d / H$ using learned linear projections. For each head, volumetric multimodal attention is computed independently following the formulation introduced in the previous section, operating on head-specific query, key, and value representations. M-VMA remains matematically valid for a number of conditioning modalities $M+1 \leq d_h$.



\begin{wrapfigure}{r}{0.5\textwidth}
    \vspace{-0.80cm}
    \centering
    \includegraphics[width=\linewidth]{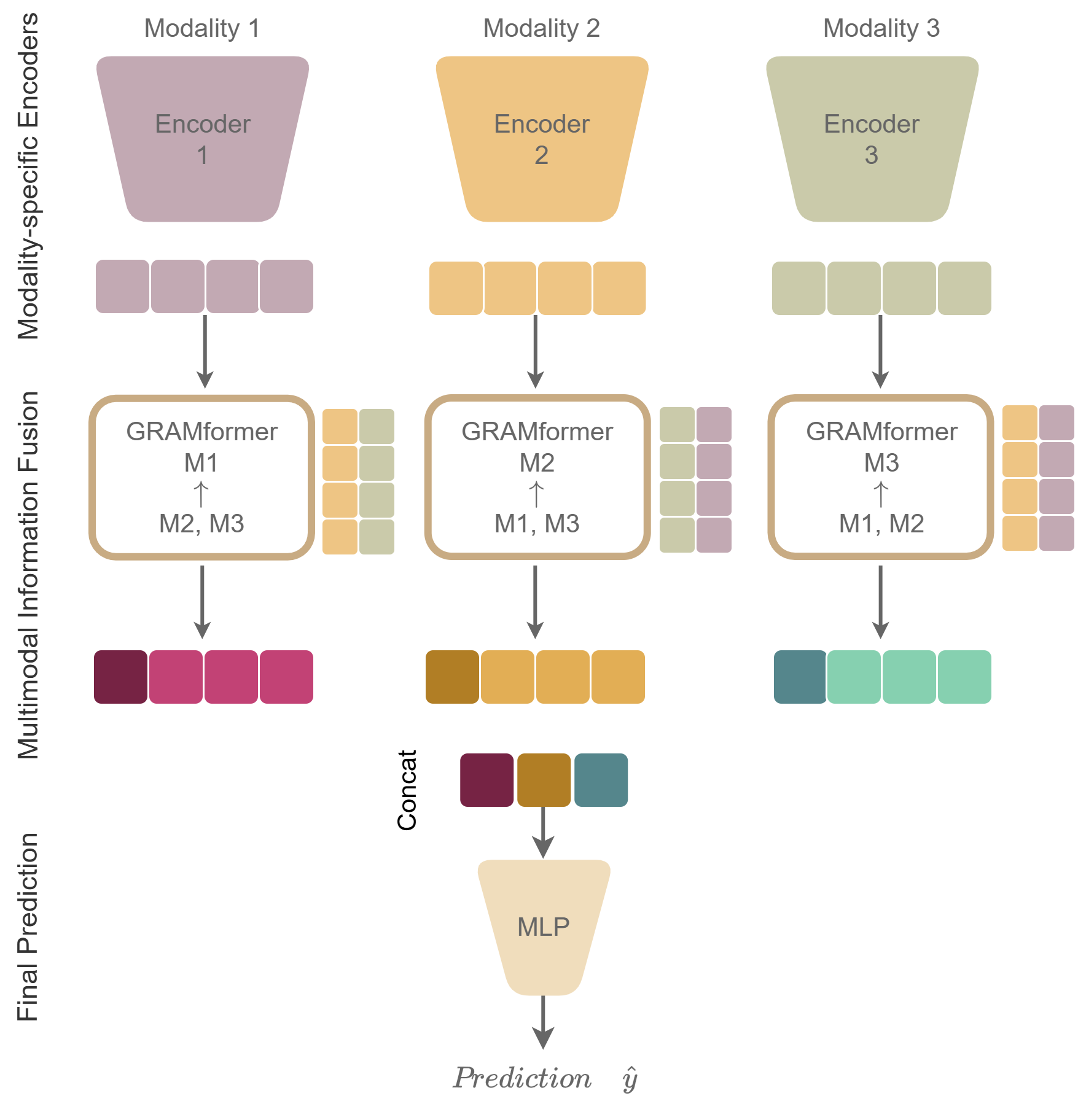}
    \caption{Proposed architecture with GRAMformer for multimodal information fusion.}
    \label{fig:gramformer}
    \vspace{-1.0cm}
\end{wrapfigure}
\section{GRAMformer}

The proposed GRAMformer is a multimodal transformer architecture that builds directly upon the standard transformer encoder design, with a single yet crucial modification: conventional cross-attention layers are replaced with the proposed Volumetric Multimodal cross-Attention (VMA) or its multi-head extension (M-VMA). Aside from this substitution, GRAMformer preserves the architectural principles, inductive biases, and training procedures of classical transformer-based models.
Concretely, each GRAMformer layer follows the familiar structure of a multimodal cross-attention block implemented via VMA or M-VMA, and a position-wise feed-forward network, with residual connections and layer normalization applied throughout. This design ensures that GRAMformer remains fully compatible with existing transformer implementations while enabling higher-order multimodal interactions.
Crucially, leveraging multiple modalities via VMA enables the GRAMformer to capture multimodal interactions without requiring additional network parameters. Therefore, the proposed GRAMformer remains lightweight as modalities scale up while improving the performance in downstream tasks.

\subsection{Multimodal Classification Architecture}
Figure~\ref{fig:gramformer} illustrates how GRAMformer is instantiated for multimodal classification tasks. The model processes multiple temporally aligned input modalities (for instance, text, audio, and visual streams), each of which is first encoded independently using modality-specific encoders. These encoders may be frozen or finetuned depending on the experimental setting.
The resulting unimodal token sequences are then fused through a stack of GRAMformer layers. In this fusion stage, in turn, each modality (typically text, as usually proven to be relevant for multimodal tasks \cite{Zhu2023LanguageBindEV}) is designated as the query stream, while the remaining modalities act as conditioning modalities. Multimodal fusion is performed through the VMA/M-VMA cross-attention block, which jointly attends to all conditioning modalities at each layer, explicitly modeling their interactions rather than relying on pairwise fusion.
After multimodal fusion, the output representations corresponding to the different modalities are aggregated by concatenating the corresponding classification tokens along the token dimensionality and passed to a classification head for final prediction $\hat{y}$.


\section{Experimental Evidences} 

\begin{table*}[]
\centering
\caption{Results in other MultiBench datasets. Params represents the number of parameters of each method. Binary accuracy for UR-FUNNY and MUsTARD; MSE for MuJoCo Push; Acc for Vision and Touch.}
\label{tab:funny}
\begin{tabular}{@{}l|c|c|c|c|c@{}}
\toprule
Method & Params & UR-FUNNY $\uparrow$ & MUsTARD $\uparrow$ & MuJoCo Push $\downarrow$ & Vision \& Touch $\uparrow$
\\ \midrule
Unimodal (T)   &   1.93 M   &    60.28        &  54.34     & -  & - \\
Unimodal (V)   &   2.05 M   &   52.64       &   54.92    & - & - \\
Unimodal (A)   &   1.53 M    &     49.33     &   54.49    & - & - \\
TF   &  5.55 M     &     61.62     &    60.00   &  0.711 & 91.57 \\
MulT        &  3.08 M   &  61.73   &  61.20 & 0.463 & 92.00 \\ \midrule
GRAMformer  &  \textbf{0.37 M}  &  \textbf{62.46}   &  \textbf{63.96}  & \textbf{0.364} & \textbf{92.90} \\ \bottomrule
\end{tabular}
\end{table*}

\begin{figure*}[t]
    \centering
    \includegraphics[width=\linewidth]{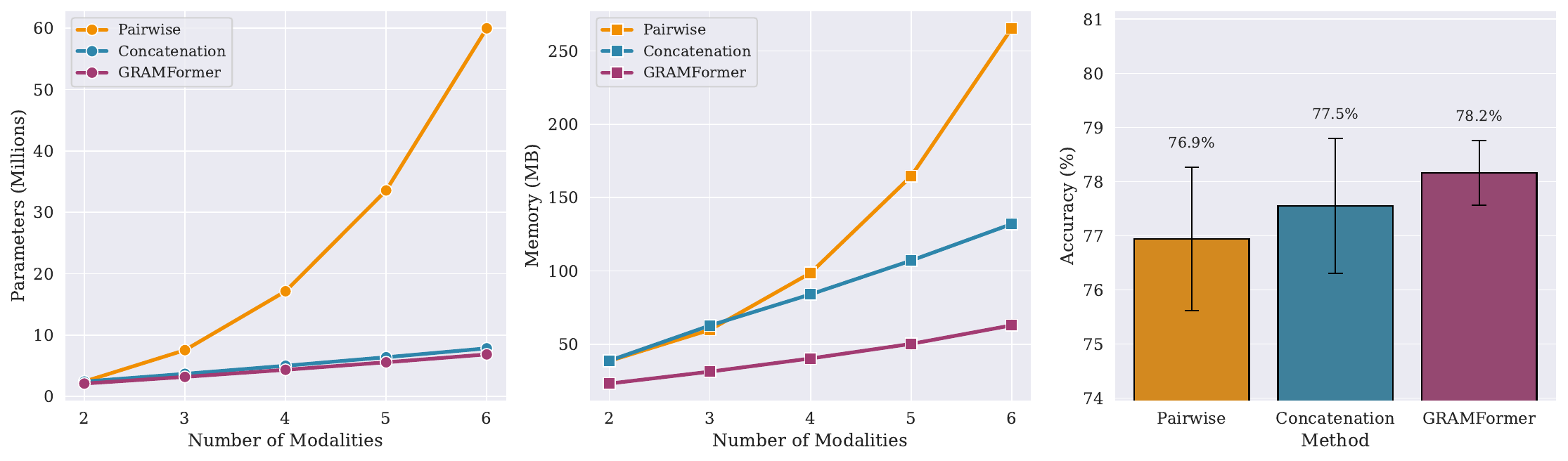}
    \caption{Number of parameters, memory usage, and accuracy scaling the number of modalities for pairwise, concatenation, and the proposed GRAMformer with VMA. Accuracy at 2 is computed on the MOSI dataset, considering the neutral class.
    }
    \label{fig:ablation_modalities}
\end{figure*}

We evaluate the proposed Volumetric Multimodal Cross-Attention (VMA) mechanism on temporally aligned multimodal sentiment analysis benchmarks to prove VMA ability to model joint cross-modal interactions beyond pairwise attention, and its scalability to more conditioning modalities.

\subsection{Setting}

\textbf{Datasets.} To be consistent with the literature \cite{Tsai2019MUlT, dufumier2025what, cheng2021EMNLP, Zheng2025MMAsia}, we conduct experiments on the MultiBench benchmark \cite{liang2021multibenchmultiscalebenchmarksmultimodal}, spanning different domains and scenarios. Specifically, we select the most common MOSI and MOSEI datasets, together with UR-FUNNY, MUsTARD, MuJoCo Push, and Vision \& Touch. MOSI and MOSEI are popular benchmarks for multimodal sentiment analysis and affective computing. UR-FUNNY and MUsTARD are benchmarks for humor and sarcasm detection, while MuJoCo Push and Vision \& Touch are robotics datasets. The affective computing and sarcasm datasets provide temporally aligned text, audio, and visual modalities. Different from the others, MuJoCo Push and Vision \& Touch are four-modal datasets and include two visual modalities (RGB and depth map) and two sensors, namely force and proprioception. Following standard practice, we treat sentiment prediction as a regression task and report results under multiple evaluation metrics obtained via thresholding the regression outputs. For MOSI and MOSEI, we consider binary sentiment classification (Acc-2), five-class classification (Acc-5), and seven-class classification (Acc-7) settings, evaluating the regression through MAE and correlation (Corr). This evaluation setup allows us to assess both fine-grained sentiment modeling and coarse polarity prediction. For UR-FUNNY and MUsTARD we just report binary classification. For MuJoCo Push, following the conventional approach for the regression task in this dataset, we employ the Mean Squared Error (MSE), while for Vision \& Touch we report the accuracy.


\begin{table*}
\centering
\caption{Performance on the MOSI dataset (average over 5 seeds). Results are reported as mean $\pm$ standard deviation.}
\label{tab:mosi_results}
\resizebox{\linewidth}{!}{
\begin{tabular}{lccccccccc}
\toprule
Model &
Acc-2 (w/ 0) $\uparrow$ &
F1-2(w/ 0) $\uparrow$ &
Acc-2 (w/o 0) $\uparrow$ &
F1-2 (w/o 0) $\uparrow$ &
Acc-5 $\uparrow$ &
Acc-7 $\uparrow$ &
MAE $\downarrow$ &
Corr $\uparrow$ &
Params \\
\midrule

\multicolumn{10}{l}{Pre-extracted Features (Text: BERT, Audio/Video Frozen)} \\
\midrule
LF-DNN        & 77.35$\pm$0.72 & 77.25$\pm$0.72 & 78.63$\pm$0.79 & 78.60$\pm$0.76 & 37.99$\pm$1.78 & 34.49$\pm$0.76 & 0.957$\pm$0.022 & 65.66$\pm$1.22 & 0.60M \\
TFN           & 77.99$\pm$0.94 & 77.92$\pm$0.81 & 79.21$\pm$1.12 & 79.21$\pm$0.96 & 39.10$\pm$1.80 & 34.20$\pm$1.76 & 0.961$\pm$0.026 & 65.75$\pm$0.94 & 9.50M \\
LMF           & 77.43$\pm$1.28 & 77.40$\pm$1.17 & 78.66$\pm$1.69 & 78.70$\pm$1.59 & 38.89$\pm$2.50 & 34.72$\pm$1.88 & 0.952$\pm$0.034 & 65.69$\pm$2.11 & 0.51M \\
MFN           & 77.08$\pm$1.36 & 77.01$\pm$1.25 & 78.38$\pm$1.51 & 78.38$\pm$1.39 & 41.84$\pm$2.24 & 36.36$\pm$1.39 & 0.946$\pm$0.025 & 65.76$\pm$1.87 & 2.17M \\
Graph-MFN     & 78.11$\pm$0.86 & 77.99$\pm$0.85 & 79.51$\pm$0.81 & 79.46$\pm$0.78 & 36.79$\pm$2.07 & 32.95$\pm$1.39 & 0.971$\pm$0.025 & 64.85$\pm$1.25 & 2.11M \\
MULT          & 78.28$\pm$0.81 & 78.29$\pm$0.78 & 79.63$\pm$0.97 & 79.71$\pm$0.94 & 41.31$\pm$1.16 & 35.48$\pm$1.08 & 0.923$\pm$0.012 & 68.34$\pm$1.03 & 2.57M \\
GRAMformer    & 78.16$\pm$0.60 & 78.12$\pm$0.58 & 79.36$\pm$0.66 & 79.37$\pm$0.63 & 38.31$\pm$1.58 & 33.50$\pm$0.74 & 0.960$\pm$0.012 & 66.41$\pm$0.52 & 0.66M \\
\midrule

\multicolumn{10}{l}{BERT Finetuning} \\
\midrule
TETFN         & 82.22$\pm$0.29 & 82.17$\pm$0.31 & 83.93$\pm$0.32 & 83.95$\pm$0.29 & 51.92$\pm$1.68 & 45.31$\pm$1.20 & 0.728$\pm$0.011 & 78.96$\pm$0.63 & 110.7M \\
Self-MM       & 82.71$\pm$0.33 & 82.66$\pm$0.31 & 84.48$\pm$0.50 & 84.48$\pm$0.47 & 53.29$\pm$0.74 & 46.18$\pm$1.23 & 0.726$\pm$0.003 & 78.57$\pm$0.72 & 109.6M \\
CENet         & 81.69$\pm$0.50 & 81.68$\pm$0.51 & 83.35$\pm$0.61 & 83.39$\pm$0.60 & 48.89$\pm$1.47 & 43.00$\pm$0.98 & 0.758$\pm$0.010 & 78.50$\pm$0.42 & 116.0M \\
MISA          & 81.49$\pm$1.02 & 81.46$\pm$0.97 & 83.02$\pm$1.38 & 83.06$\pm$1.33 & 47.32$\pm$1.90 & 41.46$\pm$1.67 & 0.777$\pm$0.033 & 77.50$\pm$0.32 & 110.6M \\
GRAMformer    & 82.86$\pm$0.49 & 82.80$\pm$0.48 & 84.54$\pm$0.55 & 84.54$\pm$0.54 & 50.61$\pm$2.03 & 44.17$\pm$1.98 & 0.732$\pm$0.017 & 78.95$\pm$0.68 & 110.1M \\
\midrule

\multicolumn{10}{l}{RoBERTa + Data2Vec Finetuning (Video Features Frozen)} \\
\midrule
MMML          & 85.91 & 85.85 & 88.16 & 88.15 & -- & 48.25 & 0.643 & 0.838 & 889.6M \\
GRAMformer    & \textbf{87.90}$\pm$0.52 & \textbf{87.86}$\pm$0.52 & \textbf{89.94}$\pm$0.70 & \textbf{89.94}$\pm$0.69 & \textbf{57.87}$\pm$2.85 & \textbf{50.05}$\pm$2.33 & \textbf{0.593}$\pm$0.020 & \textbf{0.871}$\pm$0.006 & 731.6M \\
\bottomrule
\end{tabular}
}
\end{table*}

\begin{table*}
\centering
\caption{Performance on the MOSEI dataset (average over 5 seeds). Results are reported as mean $\pm$ standard deviation}
\label{tab:mosei_full}
\resizebox{\textwidth}{!}{
\begin{tabular}{l|cccccc|cc|c}
\toprule
\textbf{Model} &
Acc-2 (w/ 0) &
F1-2 (w/ 0) &
Acc-2 (w/o 0) &
F1-2 (w/o 0) &
Acc-5 &
Acc-7 &
MAE $\downarrow$ &
Corr $\uparrow$ &
\#Params \\ 
\midrule

\multicolumn{10}{c}{Pre-extracted Features (BERT text encoder)} \\
\midrule
LF-DNN       & 78.07$\pm$4.61 & 78.57$\pm$3.86 & 82.15$\pm$1.07 & 82.00$\pm$0.87 & 53.54$\pm$0.70 & 52.26$\pm$0.59 & 0.5630$\pm$0.0042 & 73.17$\pm$0.64 & 0.57M \\
TFN          & 76.58$\pm$4.47 & 77.25$\pm$3.80 & 81.47$\pm$1.00 & 81.41$\pm$0.82 & 53.25$\pm$0.50 & 51.81$\pm$0.60 & 0.5722$\pm$0.0027 & 71.95$\pm$0.46 & 5.04M \\
LMF          & 81.30$\pm$0.70 & 81.67$\pm$0.65 & 84.28$\pm$0.37 & 84.16$\pm$0.35 & 53.96$\pm$0.29 & 52.57$\pm$0.27 & 0.5607$\pm$0.0037 & 73.51$\pm$0.48 & 0.51M \\
MFN          & 80.20$\pm$1.00 & 80.70$\pm$0.87 & 83.59$\pm$0.28 & 83.54$\pm$0.26 & 52.80$\pm$0.49 & 51.40$\pm$0.50 & 0.5706$\pm$0.0045 & 71.87$\pm$0.61 & 127.7M \\
Graph-MFN    & 82.05$\pm$0.90 & 82.25$\pm$0.67 & 84.02$\pm$0.29 & 83.81$\pm$0.44 & 53.17$\pm$0.34 & 51.86$\pm$0.31 & 0.5664$\pm$0.0018 & 72.61$\pm$0.20 & 0.68M \\
MULT         & 80.52$\pm$1.42 & 80.94$\pm$1.18 & 83.77$\pm$0.38 & 83.65$\pm$0.37 & 54.04$\pm$0.39 & 52.59$\pm$0.38 & 0.5637$\pm$0.0057 & 73.03$\pm$0.71 & 0.98M \\
GRAMformer
             & 81.49$\pm$1.25 & 81.71$\pm$1.02 & 83.55$\pm$0.35 & 83.33$\pm$0.34 & 52.78$\pm$0.45 & 51.42$\pm$0.39 & 0.5822$\pm$0.0031 & 70.84$\pm$0.48 & 0.52M \\
\midrule

\multicolumn{10}{c}{BERT Finetuning} \\
\midrule
TeTFN        & 81.44$\pm$1.81 & 81.92$\pm$1.58 & 85.15$\pm$0.64 & 85.11$\pm$0.55 & 55.82$\pm$0.75 & 53.90$\pm$0.78 & 0.5413$\pm$0.0072 & 76.12$\pm$0.43 & 110.7M \\
Self-MM      & 80.19$\pm$2.90 & 80.75$\pm$2.54 & 84.16$\pm$0.80 & 84.14$\pm$0.66 & 55.91$\pm$0.37 & 54.14$\pm$0.33 & \textbf{0.5287$\pm$0.0047} & 76.43$\pm$0.45 & 109.6M \\
GRAMformer
             & 80.97$\pm$1.25 & 81.48$\pm$1.06 & 84.89$\pm$0.28 & 84.85$\pm$0.24 & 54.78$\pm$0.44 & 53.02$\pm$0.47 & 0.5479$\pm$0.0057 & 76.10$\pm$0.18 & 110.0M \\
\midrule

\multicolumn{10}{c}{RoBERTa-Large + Data2Vec (Video features pre-extracted)} \\
\midrule
MMML         & 86.32 & 86.23 & 86.73 & 86.49 & 57.32 & 54.95 & 0.5170 & 0.790 & 889.6M \\
GRAMformer
             & \textbf{84.07} & \textbf{84.33} & \textbf{87.39} & \textbf{87.26} & \textbf{58.07} & \textbf{56.05} & \textbf{0.4963} & \textbf{0.810} & 731.6M \\
\bottomrule
\end{tabular}
}
\end{table*}

\textbf{Baselines.} We compare our approach against a diverse set of multimodal baselines that rely on pairwise fusion mechanisms.
Among all the methods, to be consistent with previous literature, we consider tensor fusion methods like TFN \cite{Zadeh2017TensorFN} and LMF \cite{Liu2018EfficientLM}, modality factorization methods as MISA \cite{Hazarika2020MISA} and Self-MM \cite{Yu_Xu_Yuan_Wu_2021}, and methods that compute pairwise dot products or concatenation among modalities as MulT \cite{Tsai2019MUlT}, MAG-BERT \cite{rahman-etal-2020-integrating}, and MMML \cite{Wu2023MultimodalMF}. As discussed in Section~\ref{sec:limitations}, such strategies are limited to pairwise interactions and cannot explicitly model joint multimodal alignment. Instead, SPT \cite{cheng2021EMNLP}, MMT \cite{ijcai2022p480}, and FDMA \cite{Zheng2025MMAsia} integrate triplet or factorized attention mechanisms to fuse multiple modalities. 


\textbf{Model architecture and training.}
Our model follows a modular multimodal transformer design. Modality-specific encoders are used to extract unimodal representations, which are then fused through a cross-modal fusion block based on the proposed Volumetric Multimodal Cross-Attention mechanism. We experiment with different encoder configurations. In one setting, unimodal encoders operate on pre-extracted features from the repository in \cite{mao2022m} and remain frozen during training. In more advanced configurations, we finetune the text encoder (BERT or RoBERTa) and, in the most expressive setting, jointly finetune both the text encoder and the audio encoder (Data2Vec). The proposed VMA module is used exclusively within the cross-modal fusion block, replacing standard cross-attention layers. Results for such different configurations are visually shown in Fig. \ref{fig:fig1} and numerically in Tab.~\ref{tab:mosi_results}.
All models are trained using comparable optimization settings, including learning rates, batch sizes, and numbers of training epochs. The overall training is kept as consistent as possible across methods. Detailed hyperparameters and training configurations are provided in Appendix A.

\subsection{Results and Discussion}

Table~\ref{tab:mosi_mosei} reports results in the MOSI and MOSEI datasets. Across all evaluation metrics, our GRAMformer with the proposed VMA consistently outperforms standard cross-attention baselines, demonstrating improved multimodal sentiment prediction.
Notably, gains are more pronounced in multi-class classification settings, suggesting that our VMA module better captures fine-grained multimodal sentiment cues. These results indicate that explicitly modeling joint interactions among text, audio, and visual modalities leads to more expressive multimodal representations than pairwise attention alone. In addition, Figure \ref{fig:fig1} shows results for different encoding settings in the MOSI dataset, with the proposed GRAMformer outperforming baselines while keeping the number of parameters lower due to its ability to effectively model multimodal interactions. Furthermore, as a confirmation, improved performance can also be observed in Tab.~\ref{tab:funny} for the UR-FUNNY, MUsTARD and MuJoCo Push datasets.
Overall, the experimental results demonstrate that our Volumetric Multimodal cross-Attention provides an effective mechanism for modeling higher-order multimodal interactions. By explicitly capturing joint alignment across multiple modalities within a single attention operation, VMA improves performance while remaining lightweight, making it a strong alternative to conventional pairwise cross-attention in multimodal transformers.

\begin{table}[t]
\centering
\caption{Ablation study on the number of conditioning modalities on the MOSI dataset. Acc-$k$ denotes classification accuracy after thresholding the regression output into $k$ classes.}
\label{tab:modalities}
\begin{tabular}{lcccc}
\toprule
Method & \# Modalities & Acc-2 $\uparrow$ & Acc-5 $\uparrow$  & MAE $\downarrow$ \\
\midrule
Unimodal & 1 (T)                  &  84.89  &  54.81   &  0.666  \\
GRAMformer & 2 (T+A)          &  85.91  &  56.08   &  0.643 \\
GRAMformer & 3 (T+A+V)  & \textbf{87.90}  &  57.87    &  0.593  \\
GRAMformer & 4 (T+A+V+VED) &    86.73  &  \textbf{59.13}   & \textbf{0.588} \\
\bottomrule
\end{tabular}
\end{table}

\subsubsection{Scalability to Any Order of Modalities}

We evaluate the ability of the proposed VMA mechanism to naturally scale to an arbitrary number of conditioning modalities by jointly processing all modalities within a single attention operation. To this end, we conduct an ablation study on the MOSI dataset in which the number of modalities is progressively increased. The corresponding results are reported in Table~\ref{tab:modalities}. We begin with a unimodal setting using only textual inputs encoded with RoBERTa. We then incrementally incorporate additional modalities, first adding the audio modality encoded using Data2Vec. For the visual modality, we employ pre-extracted visual features obtained with OpenFace. As a fourth modality, we introduce Visual Emotional Descriptors (VED) following \cite{wu2025enriching}. Specifically, we extract facial Action Units (AUs) using OpenFace and treat them as textual descriptors, which are subsequently encoded using the same text encoder to produce VED tokens. These tokens are then incorporated into the multimodal fusion process as an additional modality.
Results show that GRAMformer effectively integrates an increasing number of heterogeneous modalities, with each additional modality contributing complementary information that is successfully exploited by the VMA modules. Performance consistently improves as more modalities are added, highlighting the ability of volumetric attention to capture higher-order multimodal interactions without architectural modifications.

Furthermore, we conduct an additional experiment to stress-test scalability by increasing the number of conditioning modalities up to six. Figure~\ref{fig:ablation_modalities} illustrates the resulting scaling behavior in terms of parameter count and memory consumption. Despite explicitly modeling higher-order multimodal interactions, GRAMformer remains lightweight and computationally efficient, exhibiting favorable scaling properties compared to pairwise fusion strategies while achieving superior predictive performance as measured by classification accuracy.

\begin{table}[]
\centering
\caption{Ablation study in the MOSI dataset, including the neutral class (w/ 0) or not (w/o 0). Gating defined in \eqref{eq:gating}, Dot Prod is dot product regularization.}
\label{tab:ablation}
\begin{tabular}{@{}ccc|ccc@{}}
\toprule
Gating & Dot Prod & $\beta$ & Acc-2 (w/ 0) & Acc-2 (w/o 0) \\ \midrule
\ding{51}   &\ding{51}&  0.0  & 77.17  & 78.32    \\
\ding{51}   &\ding{51}&  0.5  & 77.87  & 78.99   \\
\ding{51}   &\ding{51}&  1.0  & 77.37  & 78.60    \\
\ding{51}   &\ding{55}&  1.5  & 77.46  & 78.44  \\
\ding{55}   &\ding{51}&  1.5  & 76.94  & 77.99  \\ \midrule
\ding{51}   &\ding{51}&  1.5  &  \textbf{78.16} & \textbf{79.36} \\ \bottomrule
\end{tabular}
\end{table}

\subsubsection{Experiments on different encoding configurations}
We conduct extensive experiments on MOSI and MOSEI datasets within different settings and encoding configurations. In the first configuration, we use pre-extracted textual, visual, and audio features using the MMSA GitHub codebase \cite{mao2022m}. In the second configuration, we let the text encoder (BERT) be finetuned jointly with the fusion module. Finally, the third configuration comprises larger-scale experiments by finetuning RoBERTa as the text encoder and Data2Vec as the audio encoder. Results are shown in Tab~\ref{tab:mosi_results} and Tab~\ref{tab:mosei_full}. The results from Tab~\ref{tab:mosi_results} are used to build Fig~\ref{fig:fig1}. All the results are obtained by averaging 5 runs for each model with different weight initializations.

\subsubsection{Ablation Studies}
\begin{wraptable}{r}{0.45\textwidth}
\vspace{-0.40cm}
\centering
\caption{Ablation study on temporally unaligned (shifted tokens) or misaligned semantics (masked tokens) for MOSI.}
\label{tab:abl_mask}
\begin{tabular}{l c c c}
\toprule
Masked & Shifted & Acc2 (w/ 0) & Acc2 (w/o 0) \\
\midrule
20\% & 0 & 77.26 & 78.36 \\
50\% & 0 & 77.14 & 78.26 \\
0    & 2 & 77.49 & 78.60 \\
0    & 5 & 77.81 & 78.96 \\ \midrule
0    & 0 & 78.16 & 79.36 \\
\bottomrule
\end{tabular}
\end{wraptable}

To better understand the contribution of our volumetric term in VMA, we perform several ablation studies.

In the first study, we remove the volume term from \eqref{eq:attscores} by letting $\beta$ varying in $\beta=\{0, 0.5, 1.0, 1.5\}$, with $\beta=0$ reducing the formulation to a standard pairwise cross-attention. Then, we remove the dot product regularization instead, and finally we perform a test without the gating mechanism of \eqref{eq:gating}. Results reported in Tab.~\ref{tab:ablation} show a consistent drop in performance with configurations different from the proposed one. This confirms that the geometric interactions captured by the volume play a crucial role in modeling joint multimodal alignment, beyond what can be achieved through pairwise dot-product similarity alone.

In the second study, we test the robustness of VMA to temporal unaligned or misaligned semantics. Specifically, we shift the audio tokens by 2 and by 5, and we mask tokens (audio modality again) by 20\% and 50\%, to mask their semantics. Table~\ref{tab:abl_mask} shows the results, which highlight that the GRAMformer is robust to misaligned semantics or temporally unaligned tokens, preserving the performance in more challenging scenarios.

\section{Conclusions}
In this paper, we introduced Volumetric Multimodal cross-Attention (VMA), a novel attention mechanism that overcomes a key limitation of existing multimodal transformers by defining attention scores jointly over sets of modality representations rather than relying solely on independent pairwise similarities. By leveraging the volume of the parallelotope spanned by the query and multiple modality-specific keys, VMA provides a unified geometric formulation for modeling any-order multimodal interactions while efficiently scaling with the number of conditioning modalities. We integrated VMA into the GRAMformer, a novel transformer architecture that replaces conventional cross-attention with volumetric attention without altering the standard transformer structure. Empirical results on multimodal benchmarks demonstrate that this design improves performance while remaining lightweight compared to pairwise and concatenation-based fusion strategies.

\balance
\bibliography{GRAMformer}
\bibliographystyle{ieeetr}

\newpage
\onecolumn

\section*{Appendix A}
\label{sec:appendix}
\begin{table*}[h]
\centering
\small
\setlength{\tabcolsep}{4pt}
\caption{Hyperparameters across all datasets and training settings.}
\resizebox{\textwidth}{!}{%
\begin{tabular}{lcccccccc}
\toprule
Hyperparameter 
& MOSI 
& MOSEI 
& BERT 
& RoBERTa / Data2Vec 
& R-FUNNY 
& MUsTARD 
& MuJoCo Push 
& Vision \& Touch \\
& (pre-extr.) 
& (pre-extr.) 
& (FT) 
& (FT) 
& (pre-extr.) 
& (pre-extr.) 
& (pre-extr.) 
& (pre-extr.) \\
\midrule
Pretrained model            & - & - & bert-base-uncased & pretrained & - & - & - & -\\
Optimizer                   & Adam & Adam & Adam & AdamW & Adam & Adam & Adam & Adam\\
Epochs (max)                & 100 & 20 & 100 / 20 & 100 & 100 & 100 & 20 & 20 \\
Early stopping patience     & 6 & 6 & 6 & 4 & 10 & 10 & 10 & 10 \\
Batch size                  & 32 & 32 & 32 & 16 & 32 & 32 & 32 & 32\\
Learning rate               & 1e$^{-3}$ & 1e$^{-3}$ & 5e$^{-5}$ & 1e$^{-5}$ & 1e$^{-3}$ & 1e$^{-3}$ & 1e$^{-5}$ & 1e$^{-5}$ \\
Number of levels            & 6 & 4 & 6 & 4 & 6 & 6 & 3 & 3 \\
Feature dim / heads         & [40, 10] & [40, 10] & [40, 10] & [1024, 8] & [40, 10] & [40, 10] & [64, 4] & [64, 4] \\
Conv1D kernel size (L/A/V)  & 5 / 5 / 5 & 5 / 5 / 5 & - & - & 5 / 5 / 5 & 5 / 5 / 5 & - & - \\
ReLU dropout                & - & - & 0.0 & 0.1 & - & - & - & - \\
Embedding dropout           & - & - & 0.0 & 0.1 & - & - & - & - \\
Text dropout                & 0.1 & 0.3 & 0.1 & 0.1 & - & - & - & - \\
Attention dropout           & 0.1 & 0.2 & 0.1 & 0.1 & 0.1 & 0.1 & 0.1 & 0.1 \\
Audio attention dropout     & 0.2 & 0.1 & - & - & - & - & - & - \\
Visual attention dropout    & 0.2 & 0.1 & - & - & - & - & - & - \\
Attention dropout (A/V)     & - & - & - & - & 0.2 / 0.2 & 0.2 / 0.2 & - & - \\
Residual dropout            & 0.1 & 0.1 & 0.1 & 0.1 & 0.1 & 0.1 & 0.1 & 0.1 \\
Output dropout              & 0.2 & 0.2 & 0.2 & 0.1 & 0.2 & 0.2 & 0.2 & 0.2 \\
Gradient clipping           & 0.8 & 0.8 & 0.8 & 0.8 & 0.8 & 0.8 & 0.8 & 0.8 \\
Weight decay                & 1e$^{-3}$ & 1e$^{-3}$ & 1e$^{-3}$ & 1e$^{-3}$ & 1e$^{-3}$ & 1e$^{-3}$ & 1e$^{-3}$ & 1e$^{-3}$ \\
\bottomrule
\end{tabular}%
}
\label{tab:all-hparams}
\end{table*}

\end{document}